\documentclass[runningheads]{llncs}
\usepackage{graphicx}
\usepackage{amsmath,amssymb} 
\usepackage{color}
\usepackage{multirow}
\usepackage{algorithm}
\usepackage{algpseudocode}
\usepackage{subcaption}
\usepackage{arydshln}
\usepackage{float}
\usepackage{soul}
\usepackage{xcolor}

\newcommand{\etal}{\textit{et al.}}

\newcommand{\ub}{\text{UB}}
\newcommand{\lb}{\text{LB}}
\newcommand{\level}{l}
\newcommand{\coloring}{c}

\newcommand{\cM}{\mathcal{M}}
\newcommand{\cN}{\mathcal{N}}
\newcommand{\cI}{\mathcal{I}}
\newcommand{\cJ}{\mathcal{J}}

\newcommand{\cS}{\mathcal{S}}

\newcommand{\cP}{\mathcal{P}}

\newcommand{\bR}{\mathbf{R}}
\newcommand{\bt}{\mathbf{t}}

\newcommand{\bbR}{\mathbb{R}}
\newcommand{\cC}{\mathcal{C}}

\newcommand{\bm}{\mathbf{m}}
\newcommand{\bn}{\mathbf{n}}
\newcommand{\cG}{\mathcal{G}}
\newcommand{\cV}{\mathcal{V}}
\newcommand{\cE}{\mathcal{E}}

\newcommand{\NumberOfSemanticClasses}{$L$~}
\newcommand{\algo}

\begin{document}
\pagestyle{headings}
\mainmatter
\def\ECCV18SubNumber{6149}  

\title{Fast Semantic-Assisted Outlier Removal for Large-scale Point Cloud Registration} 


\author{Giang Truong\inst{1} \and
Huu Le\inst{2} \and
Alvaro Parra\inst{3} \and 
Syed Zulqarnain Gilani\inst{1} \and
Syed M. S. Islam\inst{1} \and
David Suter\inst{1}}
\authorrunning{G. Truong et al.}
%
\institute{School of Science, Edith Cowan University, Australia \and
Department of Electrical Engineering, Chalmers University of Technology \and
School of Computer Science, The University of Adelaide, Australia\\}

\maketitle

\begin{abstract}
With current trends in sensors (cheaper, more volume of data) and applications (increasing affordability for new tasks, new ideas in what 3D data could be useful for); there is corresponding increasing interest in the ability to automatically, reliably, and cheaply, register together individual point clouds.  The volume of data to handle, and still elusive need to have the registration occur fully reliably and fully automatically, mean there is a need to innovate further. One largely untapped area of innovation is that of exploiting the {\em semantic information} of the points in question. Points on a tree should match points on a tree, for example, and not points on car. Moreover, such a natural restriction is clearly human-like - a human would generally quickly eliminate candidate regions for matching based on semantics. Employing semantic information is not only efficient but natural. It is also timely - due to the recent advances in semantic classification capabilities. This paper advances this theme by demonstrating that state of the art registration techniques, in particular ones that rely on ``preservation of length under rigid motion'' as an underlying matching consistency constraint, can be augmented with semantic information. Semantic identity is of course also preserved under rigid-motion, but also under wider motions present in a scene.
We demonstrate that not only the potential obstacle of cost of semantic segmentation, and the potential obstacle of unreliability of semantic segmentation; are both no impediment to achieving both speed and accuracy in fully automatic registration of large scale point clouds. In particular, we take a state of the art approach, that uses an (approximate) maximum clique heuristic (a clique on edges encapsulating matches that preserve length), to yield a version further enhanced with exploitation of semantic information. In essence, cheaply and reliably (for the purposes required) pruning the data that has to be dealt with. 

\end{abstract}

\section{Introduction}
Point cloud registration is an important problem underpinning a wide range of computer vision applications: such as object recognition~\cite{chua1997point}, real-time depth fusion~\cite{newcombe2011kinectfusion} and localization~\cite{elbaz20173d} among others. Given two sets of points in three-dimensional (3D) space,
the goal is to estimate a rigid-body transformation, consisting of a rotation matrix $\bR \in SO(3)$ and a translation vector $\bt \in \bbR^3$, that best aligns the two input point sets. While a variety of methods exist in the literature to address this problem~\cite{fischler1981random,parra19pairwise,bergstrom2014robust}, handling a large proportion of erroneous matches (outliers) present in many large scale datasets, is still a difficult challenge for existing methods. 
To address this problem, we propose a novel method for fast outlier removal that can be used -- as a data pre-processor -- to significantly boost the performance of existing state-of-the-art algorithms.

The task of robust registration is typically done by employing randomized strategies, where RANdom Sample Consensus (RANSAC)~\cite{fischler1981random} and its variants~\cite{torr2000mlesac,chum2003locally,chum2005matching} are often the tools of choice. RANSAC is an appealing choice due to its simplicity, while providing relatively satisfactory results in problem instances having low outlier ratios. 
However, as the number of outliers grows, RANSAC can no longer sustain its performance, since it may require an exponentially large number of iterations to produce acceptable results~\cite{fischler1981random}. This in turn can prove to be a bottle-neck for real-time applications. 

To achieve a high-quality estimate, within a reasonable amount of run-time, it is beneficial if a fraction of outliers could be removed before executing the core robust fitting algorithms. Along this line of research, several approximate or exact methods have been developed for the outlier removal task, e.g.,~\cite{parra2015guaranteed,gore6dof,Svarm2014,Giang2019}. However, such methods require the evaluation for every single pair in the initial putative correspondence set, where each pair requires the execution of a costly branch-and-bound sub-problem. 
Note that in large-scale applications, 
methods such as \cite{gore6dof} are impractical. 

The motivation behind our work comes from human vision, where the reasoning process is guided by 
semantic information.
For example, we would only attempt to match key points that belong to the same semantic class (e.g., houses, trees, cars, etc.).
In recent years, semantic information for 2D images or 3D point clouds can also be easily obtained in real-time. However, how to effectively utilize the available semantic information to support classical robust registration algorithms remains an interesting research question. 

In this work, we propose novel techniques to partially address the above question. 
Specifically, we investigate a novel combination between traditional robust estimation techniques -- in particular finding pairwise consistent correspondences using maximum clique~\cite{parra19pairwise}-- with semantic segmentation, which can be extracted from any well-known semantic segmentation framework \cite{wu2017squeezeseg,wu2018squeezesegv2,Darknet53}. 
Maximum clique an ingredient of choice in our work, due to the fact that it offers a convenient mechanism to exploit the available semantic information to massively reduce the problem size. Moreover, the graph formulation underlying maximum clique allows us, under the guidance of semantic labels, to decompose the original problem into multiple sub-problems, each of which  can be solved efficiently. 
Maximum clique has been used previously for rigid registration~\cite{pairwise,parra19pairwise}, 
since the clique extracted is a set of pair-wise consistent (under preservation of distance between points) correspondences, which can be used to estimate the optimal transformation.
However, existing works focus on solving this problem on graphs larger than is actually necessary, hence they are still impractical for large datasets. 
We show that with a novel use of semantic information, one only needs to solve MC sub-optimally and on smaller graphs.


The correspondence set generated by our method can then be processed by any state-of-the-art robust registration algorithm, but with significantly faster run-time since a large proportion of outliers have been effectively removed. Experimental results show that we achieve competitive or even better registration results, while our total run-time (semantic segmentation, outlier removal + robust registration) is of orders of magnitude faster than existing registration algorithms, including techniques that provide globally optimal results~\cite{gore6dof}.

It should be noted that our work differs from recent deep-learning approaches \cite{DeepVCP,3dfeatnet} for point cloud registrations, as the main work-horse behind our work is still the well-known classical robust registration algorithm (i.e., based on maximum clique) allowing us to achieve results that are close to globally optimal solutions. 
Moreover, fully deep learned approaches usually require the design of a new network architecture, which must be re-trained before they can be used on any new dataset. Our method, on the other hand, only requires off-the-shelf semantic classifiers that have been pre-trained. Moreover, our underlying ideas are agnostic as to whether the semantic information comes from 3D data, image data, or both.
\vspace{-5mm}
\section{Related Work}
\vspace{-3mm}
The task of robustly aligning two point sets has long been an active research topic in computer vision. 
One of the most well-known methods 
is RANSAC \cite{RANSAC}, which is based on repetitively sampling of points to generate model hypotheses 
to discover the model with the largest consensus set. 
Variants \cite{ransacfamily2009} of RANSAC seek to reduce computing time through guided sampling, and accelerating the evaluation of a hypothesis. 
Although RANSAC and its variants are simple and easy to implement, processing time, 
grows exponentially when dealing with the high percentages of outliers. 
4-Points Congruent Sets (4PCS) \cite{amo_fpcs_sig_08} and its variant \cite{THEILER2014149}, have better sampling strategies. However, such methods are sensitive to hyper-parameters, such as the approximate percentage of overlapping points, etc.

In contrast to heuristic methods, several globally optimal algorithms \cite{BnB2009,GOGMA} have been introduced.
These methods are mostly based on the branch-and-bound (BnB) technique.
Unfortunately, these methods are  computationally very expensive. 
To overcome these limitations, 
efforts have been made to remove a large number of outliers before applying global optimal algorithms (e.g. BnB). Parra et al.\cite{gore6dof} proposed Guaranteed Outlier Removal (GORE) to remove as many certain outliers as possible. 
Although GORE helps avoid using BnB on all elements of initial match set, it is still computationally expensive as it calculates complex bounding functions.
Other work \cite{pairwise,parra19pairwise} sought to quickly discover the largest set of consistent correspondences from an initial match set. Transformation parameters may be extracted easily and quickly from that subset by applying robust algorithms (e.g. RANSAC \cite{RANSAC}). 
However, these methods are still computationally expensive when dealing with large-scale datasets.

\vspace{-3mm}
\section{Background}
\vspace{-3mm}
\subsection{Problem Definition}
\vspace{-3mm}
Consider two input point clouds $\cM = \{\bm_i \in \bbR^3\}_{i=1}^M$ and $\cN = \{\bn_i \in \bbR^3\}_{i=1}^N$.
(in which the number of points in each input point set can be in the ranges of $M,N > 100,000$ points). 
As common,
we assume that a set of putative correspondences $\cP$, has been obtained by any off-the-shelf key point extraction techniques \cite{ISS,Keypoint2013}, and are given as $\cP=\{p_k\}_{k=1}^{D}$, where each correspondence pair $p_k = ( \mathbf{m}_k, \mathbf{n}_{k'} )$, out of the set of $D$ correspondences, contains $\bm_k \in \cM$ and $\bn_{k'} \in \cN$. Our goal is to estimate the optimal 6DOF rigid body transformation (i.e. $\bR^* \in SO(3)$ and $\bt^* \in \bbR^3$) that maximizes the following objective function:
\vspace{-3mm}
\begin{equation}
\label{eq:maxcon_objective}
    (\bR^*, \bt^*) = \arg\max_{\bR, \bt} \sum_{k=1}^{D} \mathbb{I}(\| \mathbf{R}\mathbf{m}_k + \mathbf{t} - \mathbf{n}_{k'} \| \leq \epsilon),
    \vspace{-3mm}
\end{equation}
where $\|\cdot\|$ denotes the $\ell_2$ Euclidean norm in $\bbR^3$. The indicator function $\mathbb{I}(\cdot)$ returns 1 if its predicate $(\cdot)$ is true and 0 otherwise. 
The maximum consensus objective function \eqref{eq:maxcon_objective} has been used widely e.g.,~\cite{gore6dof,Giang2019}. The hyper-parameter $\epsilon$ specifies the inlier threshold, which can be chosen based on prior knowledge about the problem.

Due to the noisy characteristic of input sensors and the imperfection of existing feature extraction techniques, $\cP$ may contain a large fraction of outliers (the KITTI dataset can contain up to $\rho = 90\%$ of outliers - please refer to the supplementary material). In order to estimate $\bR^*$ and $\bt^*$, a subset $|\cI^*|$ containing $(1-\rho)D$ inliers must be identified, and vice versa (i.e., if the optimal transformation $(\bR^*, \bt^*)$ is given, $\cI^*$ can easily be obtained). 
In this work, we introduce a novel approach to effectively remove outliers in $\cP$ to obtain a subset $\cP' \subset \cP$, where the outlier ratio $\rho'$ of $\cP'$ is much lower than that of the original set $\cP$. Given $\cP'$, the optimal transformation between the two point clouds can then be obtained by, e.g., running RANSAC.

In fact, our outlier removal approach is motivated by a class of registration methods that attempt to solve point cloud alignment {\em without putative correspondences} \cite{parra19pairwise,pairwise},  which uses pairwise length constraints to obtain the optimal set of correspondences by searching for the maximum clique. 
However, in practice, using all the points as input to such algorithms renders them impractical hence we still use keypoints. 

In the next section, we briefly introduce the Maximum Clique algorithm, which serves as a backbone for our algorithm.
\vspace{-5mm}
\subsection{Maximum Clique (for preserved distance between point pairs) Formulation}
\label{sec:max_clique_formulation}
\vspace{-3mm}
As previously discussed, when the set $\cI^*$, containing the highest number of inliers (set with maximum consensus) is achieved, the optimal transformation $(\bR^*, \bt^*)$ can be obtained using, e.g., SVD \cite{SVD} or RANSAC\cite{RANSAC}. Following \cite{parra19pairwise,pairwise}, outliers for such an optimal subset can be removed by solving the maximum set of consistent pairs. Specifically, we first define an undirected graph $\cG=(\cV, \cE)$, where each vertex $v_k \in \cV$ corresponds to a pair of correspondences $p_k$ between the two point clouds, i.e., $\cV = \cP $. The set of edges $\cE$ of $\cG$ is defined as 
\vspace{-3mm}
\begin{equation}
    \label{eq:edge_set_definition}
    \cE = \{ (p_i, p_j) \in \cP \times \cP \,|\,\,  d (p_i, p_j) \le 2\epsilon \}
    \vspace{-2mm}
\end{equation}
where the function $d(p_i, p_j)$ specifies the distance between the two correspondence pairs, $p_i = (\bm_i, \bn_{i'})$ and $p_j = (\bm_j, \bn_{j'})$, which can be expressed as
\vspace{-2mm}
\begin{equation}
\label{eq:pairwise_distance}
d(p_i, p_j) = | \|\bm_i - \bm_j\| - \|\bn_{i'} - \bn_{j'}\|   |,
\vspace{-2mm}
\end{equation}
and $\epsilon$ is the user-defined inlier threshold. To assist further discussion in later sections, we also define a clique $\cC=(\cV', \cE'), \cV' \subseteq \cV, \cE' \subseteq \cE$ to be a sub-graph of $\cG$ 
in which there exists a connection between any two vertices. i.e., 
\vspace{-2mm}
\begin{equation}
    \label{eq:clique_edges}
    \cE' = \{(v_i, v_j) \in \cV |\, (v_i, v_j) \in \cE ,\, \forall i,j\}.
    \vspace{-2mm}
\end{equation}

For brevity, we define clique size $|\cC|$ to be the number of vertices in $\cC$. A maximum clique $\cC^*$ of $\cG$ is the clique with the maximum number of vertices. For a vertex $v_i \in \cV$, let $\Gamma(v_i)$ be a set that all elements are adjacent to $v_i$, i.e., $\Gamma(v_i) = \{ v_j \in V \,|\, (v_i, v_j) \in \cE \}$. We call $|\Gamma(v_i)|$ as the degree of vertex $v_i$.

From~\eqref{eq:pairwise_distance}, a pair of correspondences  $p_i$ and $p_j$ are said to be \emph{consistent} if $d(p_i,p_j) \le 2\epsilon$. The graph $\cG$ defined above is therefore referred to as \emph{consistency graph}. Intuitively, if $p_i$ and $p_j$ both belong to the optimal inlier set $\cI^*$, it is expected that the difference in lengths between the two segments $\overline{\bm_i \bm_j}$ and $\overline{\bn_{i'} \bn_{j'}}$ must not exceed $2\epsilon$ (in noise-free settings, $\epsilon = 0$).

The task of removing (not necessarily all) outliers  in $\cP$ can be addressed by solving for the optimal set $\cI^*$ that contains the maximum number of consistent correspondence pairs. Since every pair of inliers $i,j\in\cP$ must satisfy $d(p_i, p_j) \leq 2\epsilon$, we can remove outliers by solving
\vspace{-2mm}
\begin{equation}
\label{eq:max_clique}
    \max_{\cI \subseteq \{1,...,D\}} | \cI | \;\;
    \text{s.t.} \,\,\,\, d(p_i, p_j) \leq 2\epsilon , \forall i, j \in \cI. \\
\vspace{-2mm}
\end{equation}

Together with the introduction of the consistency graph $\cG$ as discussed above, maximizing Eq.~\eqref{eq:max_clique} is equivalent to searching for the maximum clique on $\cG$, which will be discussed in the following section. Interested readers are referred to~\cite{parra19pairwise} and supplementary material for more details.

\vspace{-5mm}
\subsection{Solving Maximum Clique}
\label{sec:solving_max_clique}
\vspace{-2mm}
Removing outliers by finding the maximum clique in the consistency graph  Eq.~\eqref{eq:max_clique} is still computationally challenging as maximum clique is a well-known NP-hard problem. The majority of optimal algorithms are based on BnB
For instance, Parra~\etal~\cite{parra19pairwise} introduced an efficient algorithm by deriving an efficient bounding function. However, such algorithms can only handle input data with modest sizes, while large-scale datasets are still considered impractical. To assist in the introduction of our algorithm and for the sake of completeness, in this section, we quickly summarize the general BnB approach to tackle Eq.~\eqref{eq:max_clique}. Our novel modification will be introduced in Section 4.

A general strategy for solving maximum clique using BnB is traversing the input graph $\cG = (\cV, \cE)$ in a depth-first-search manner. At the beginning of the algorithm, all the vertices in $\cV$ are pushed into a set $\cS$, i.e., $\cS = \cV$. A node $v_i\in \cS$ is selected and its adjacent nodes are extracted into a set $\cS'$ for further exploration. During the traversal, the optimal clique size $R^*$ obtained so far is recorded. When a node $v_i$ is considered for expansion, a bounding function $\ub(v_i)$ is computed, which provides the upper-bound for the maximum clique containing $v_i$. The computation of $\ub$ usually involves solving for $\coloring(v_i)$, which is an approximate version of graph-coloring. The upper bound is then computed as  $\ub(v_i) = \coloring(v_i) + \level(v_i)$, where $\level(v_i)$ is the current level of $v_i$ with respect to the current search tree. If $\ub(v_i) > R^*$, the search recursively continues from $v_i$ to deeper levels. Otherwise, the exploration for $v_i$ is terminated, $v_i$ is removed from $\cS'$, and the process continues from other nodes in the queue $\cS'$ until $\cS' = \emptyset$. The search is then back-tracked to investigate the remaining nodes in $\cS$. More details about BnB algorithm for maximum clique can be found in~\cite{parra19pairwise}. It is well-known that BnB has exponential complexity~\cite{horst2013global}, thus real-time applications rarely employ such techniques, unless the problem size is relatively small.

While BnB is a standard framework, deriving a tight bounding function $\tilde{R}(v)$ is crucial in determining the execution time. 
Most of the existing research relies on approximate graph-coloring to obtain the lower bound. In addition, other techniques such as ranking the nodes by their degrees, or ~\cite{parra19pairwise} can also be used to speed up the algorithms. However, to the best of our knowledge, most algorithms are still slow for large-scale problems, especially when the graph $\cG$ is dense.
\vspace{-5mm}
\section{Proposed Method}
\vspace{-3mm}
\begin{figure}[tb]
    \centering
    \includegraphics[width=1.0\linewidth]{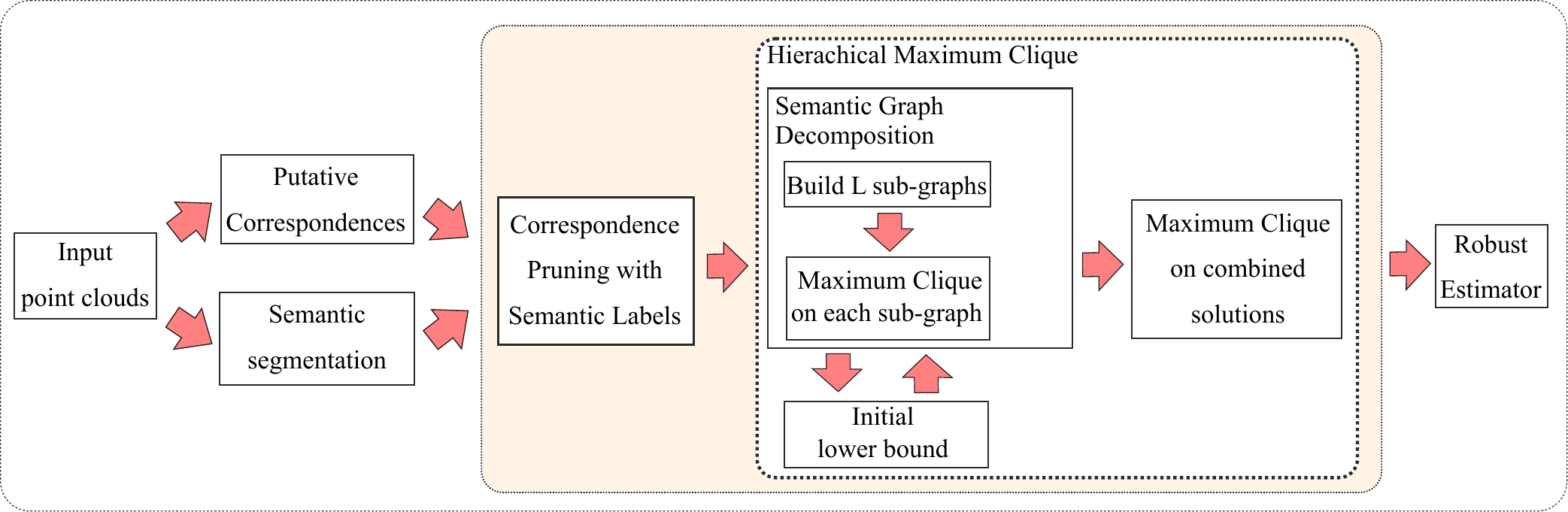}
    \caption{Overall block diagram of the proposed approach.}
    \label{fig:main_diagram}
    \vspace{-7mm}
\end{figure}

A schemetic of our proposed approach is depicted in Fig~\ref{fig:main_diagram} while the following sections show how integrating semantic information on approximating the maximum clique problem addresses the challenges discussed above. In particular, we first discuss the correspondence pruning using semantic labels. Next, using the provided semantic information containing \NumberOfSemanticClasses different classes, we propose to decompose the graph into \NumberOfSemanticClasses sub-graphs, and solve maximum clique for each of the sub-graphs. The solutions from the sub-problems are then combined in an hierarchical manner. Notably, we also demonstrate a novel use of information obtained from the solved sub-graphs to accelerate the computation of lower-bounding functions during optimizing the remaining sub-graphs. This technique allows us to achieve real-time performance on very large datasets containing an extremely high proportion of outliers. 




\vspace{-5mm}
\subsection{Correspondence Pruning with Semantic Labels}
\label{sec:CorrPruning}
\vspace{-2mm}
We use semantic labels to prune the set of input putative correspondences $\cP$. Specifically, for each 3D point $\bm \in \bbR^3$, let $l(\bm) \in \{1, \dots, L\}$ denote the semantic label (of the \NumberOfSemanticClasses semantic classes) assigned to $\bm$. Based on such a labelling, we prune $\cP$ by removing correspondences $p_k=(\bm_k,\bn_{k'})$ where $l(\bm_k) \ne l(\bn_{k'})$. Formally, after pruning, we obtain a new set $\cS \subseteq \cP$:
\vspace{-2mm}
\begin{equation}
    \label{eq:putative_after_prune}
    \cS = \{(\bm_k, \bn_{k'}) \in \cP \,|\,\, l(\bm_k) = l(\bn_{k'}) \}.
    \vspace{-2mm}
\end{equation}
  
After this pruning procedure, due to the inevitable errors caused by the underlying semantic classifiers, two problems may arise: 
\vspace{-2mm}
\begin{itemize}
    \item True inliers may be incorrectly removed, and
    \item The set $\cS$ can still contain a large fraction of outliers.
    \vspace{-2mm}
\end{itemize}
For the former problem, our empirical experiments (please refer to the supplementary material) show that our method incorrectly removes an insignificant fraction of correspondences in $\cI^*$. Moreover, this does not cause a significant impact on the final estimated transformation compared to ground-truth (See Section~\ref{sec:experiments}, for performance results).

The latter problem, on the other hand, is still challenging. One would expect that after pruning $|\cP|$, since $|\cS| < |\cP|$, applying any state-of-the-art robust fitting algorithm to $\cS$ would be trivial than the original problem of applying it to $\cP$. However, merely pruning the dataset based on semantic class matching still leaves a large number of outliers (please refer to the supplementary material). 

 
In the following section, we show how  semantic information can further be utilized to devise a fast algorithm, which can precisely remove outliers in $\cS$.

\vspace{-5mm}
\subsection{Hierarchical Maximum Clique}
\vspace{-3mm}

Approximating the maximum clique up to a constant factor has been shown to be  NP-complete~\cite{feige1991approximating}, unless P=NP. Therefore, most of the research involving maximum clique resort to heuristic mechanisms and approximate maximum clique.
Inspired by recent works on maximum clique for very large graphs~\cite{pelofske2019solving,lu2017finding,kurita2019constant}, where graph decomposition plays a major role, we propose a hierarchical maximum clique algorithm for outlier removal. 

Our heuristic algorithm stems from the special structure of rigid-body point cloud registration. Recall that, in order to roughly estimate the rotation and translation from the two given point clouds, only a subset of inliers are sufficient. The rest of the inliers can be obtained after applying the estimated $\bR$ and $\bt$. Therefore, it is adequate for an algorithm to approximate the set of inliers. 
In addition, in the context of point cloud registration, it is expected that matching points should belong to the same semantic class. Therefore, we propose to approximate this set of inliers by first using semantic information to decompose the original graph, then combine the solutions in a hierarchical manner. The algorithms are discussed in detail as follow.
%

\begin{figure}[tb]
    \centering
    \includegraphics[width=1\linewidth]{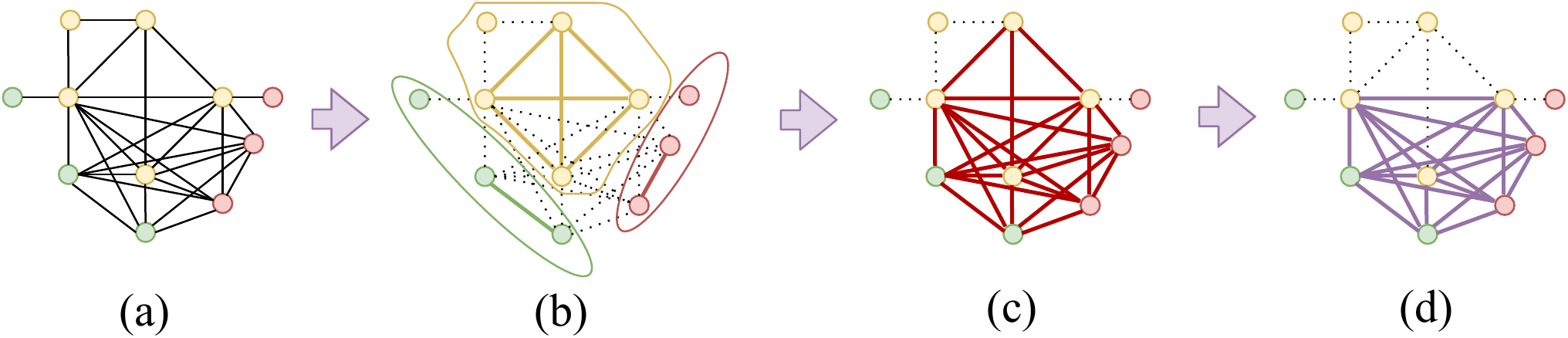}
    \caption{Illustrations of Semantic Graph Decomposition. We have three semantic classes with three different vertex colors (red, yellow and green).
    (a) The original graph with semantic information after correspondence pruning with semantic labels.
    (b) Maximum cliques of sub-graphs.
    (c) The combined maximum cliques of sub-graphs.
    (d) Final maximum clique.}
    \label{fig:graph_decomposition}
    \vspace{-7mm}
\end{figure}
\vspace{-5mm}
\subsubsection{Semantic Graph Decomposition}
After executing the semantic pruning procedure (Section~\ref{sec:CorrPruning}) to obtain the set $\cS$, let us consider the graph $\cG_S = (\cV_S, \cE_S)$ built from $\cS$ based on the technique discussed in Section~\ref{sec:max_clique_formulation}. Since the labels assigned to points in each correspondence $p_k =(\bm_k, \bn_{k'}) \in \cS$ are identical, with the abuse of notation, we also use $l(p_k)$ to denote the label associated with points belonging to $p_k$, i.e., $l(p_k) = l(\bm_k) = l(\bn_{k'})$. Also, for a graph vertex $v_k \in \cV_S$, we denote by $l(v_k)$ the label of the associated correspondence $p_k$.

Now, define \NumberOfSemanticClasses sub-graphs $\{\cG_i = (\cV_i, \cE_i)\}_{i=1}^L$ extracted from $\cG_S$, where $\cV_i$ comprises correspondences having label $i$, i.e., $\cV_i = \{ v_k \in \cV_S \,|\,\, l(v_k) = i\}$, and $\cE_i = \{ (p_p, p_q) \in \cE_S \,|\,\, l(p_p) = l(p_q) \}$. Note that by decomposing the original graph $\cG_S$ into \NumberOfSemanticClasses sub-graphs, the total number of vertices remains  the same, while some edges that connect vertices having different labels are temporarily removed. Therefore, the number of edges in each sub-graph can be massively reduced compared to the original graph, i.e., $|\cE_i| \ll |\cE_S| \;, \forall i=1,\dots\, L$, allowing us to solve maximum clique efficiently on each $\cG_i$ (see Fig.~\ref{fig:graph_decomposition} for an illustration of graph decomposition). Moreover, note that based on the sub-graph definitions, in order to obtain the sub-graphs, instead of constructing the original graph $\cG_S$ and perform the decomposition, it is sufficient to just construct $\cG_i$ based on a subset of vertices having label $i$. 

After decomposing the original graph into $L$ sub-graphs, it is natural to execute the maximum clique algorithm on the sub-graphs. For each sub-graph $\cG_i$, we denote by $\cI^*(\cG_i)$ 
the optimal inliers after solving maximum clique for graph $\cG_i$. 
Certainly, if $\cI^*(\cG)$ denotes the optimal inliers of the original graph $\cG$, it can be seen that $\bigcup_{i=1}^L\cI^*(\cG_i) \neq \cI^*(\cG)$. In other words,  combining the inlier sets obtained after solving each sub-problem does not guarantee to recover the optimal solution for the original problem. However, our empirical experiments show that combining the solutions $\bigcup_{i=1}^L\cI^*(\cG_i)$ in most  cases is sufficient to recover the optimal transformation.
\vspace{-5mm}
\subsubsection{Combining sub-optimal solutions}
After solving the sub-problems to obtain the set $\cJ  = \bigcup_{i=1}^L\cI^*(\cG_i)$, $\cJ$ may still contain outliers. One of the possible reasons is the presence of moving objects. For instance, the same car can be located at two different positions in the two input point clouds. In this scenario, points belong to such car become ``outliers" w.r.t. the remaining points. 
Moreover, the noisy set of putative correspondences may contain correspondences that are not meaningful to the underlying point cloud structure, but may still form a maximum clique. Therefore, to remove the remaining outliers in $\cJ$, the final stage in our algorithm is to solve for maximum clique on the set $\cJ$. 


\vspace{-5mm}
\subsection{Faster Tree-Pruning with Tight Initial Lower Bound}
\vspace{-2mm}
Recall that, in classical maximum clique solvers, when each vertex $v_i$ is visited, a bounding function $\ub(v_i)$ is computed (see Section~\ref{sec:solving_max_clique}), which determines if the vertex $v_i$ should be expanded. For our particular problem, we show that while solving for $\cI^*(\cG_i)$, an initial lower bound $\lb(\cG_i)$ can be used in combination with $\tilde{R}(v_i)$ to effectively prune the search tree.

We make use of the decomposed structure in our problem and utilize the solutions obtained from $\bigcup_{i=1}^{k} \cI^*(\cG_i)$ (where $k$ is the number of classes out of $L$ classes processed so far) to compute  during the process of solving for $\cI^*(\cG_i)$. 
Our new strategy is motivated by the fact that in order to obtain an acceptable transformation, only a small subset of correspondences $\hat{\cI} \subset \cI^*$, where $|\hat{\cI}| \ge 3$ is required. Therefore, after obtaining the set $\bigcup_{i=1}^{k} \cI^*(\cG_i)$, we aim to seek $(R_{i-1}, t_{i-1})$ from this subset using RANSAC. Because the fraction of outliers in $\bigcup_{i=1}^{k} \cI^*(\cG_i)$ is relatively low and $|\bigcup_{i=1}^{k} \cI^*(\cG_i)|$ is small, the run-time required by RANSAC, in most of the cases, is very fast. Then, based on $(R_{i-1}, t_{i-1})$, we compute approximate number of inliers $|I_{i}|$ and use it as initial lower bound for solving $\cI^*(\cG_i)$:
\vspace{-2mm}
\begin{equation}
    \lb(\cG_i) = |I_i| \leq |\cI^*(\cG_i)|
    \vspace{-2mm}
\end{equation} 
At any given node $v$, it is expected that $\lb(\cG_i) \le \cI^*(\cG_i) \le \ub(v)$. Consequently, at a particular node $v$, if $\ub(v) < \lb(\cG_i)$, it can be concluded that the node $v$ can be pruned.

 $|I_{i}|$ may become a tight initial lower bound because, firstly, the maximum clique may have some, but not many outliers. As a result, $|I_i|$ is quite close to $|\cI^*(\cG_i)|$. Secondly, the run-time to calculate $|I_{i}|$ is significantly fast.

\vspace{-5mm}
\begin{algorithm}[t]
\caption{Fast Semantic-Assisted Outlier Removal.}
\begin{algorithmic}[1]
\Require A set of putative correspondences $\cP$
\State Prune $\cP$ to obtain semantic consistent set $\cS = \{p_k = (\bm_k, \bn_{k'}) \in \cP | l(\bm_k) = l(\bn_{k'}) \}$.
\State Build L sub-graphs $\{\cG_i = (\cV_i, \cE_i)\}_{i=1}^L$
\For{$i = 1, ...,L$}
    \If{($i > 1$)}
        \State $\lb(\cG_i) \gets |I_i|$
    \EndIf
    \State Calculate $\cI^*(\cG_i)$ using PMC with $\lb(\cG_i)$.
    \If{($|\bigcup_{1}^i\cI^* (\cG_i)| > 3$)}
        \State Estimate ($\bR^*_i$, $\bt^*_i$) using RANSAC on $\bigcup_{1}^i\cI^* (\cG_i)$.
    \EndIf
    \State Calculate initial lower bound for next class $I_{i+1}$
\EndFor
\State Combine sub-optimal solutions from L sub-graphs $\cJ = \bigcup_{i=1}^L \cI^* (\cG_i)$
\State Run PMC again on $\cJ$ to get $\cI^*(\cJ)$
\State Estimate ($\bR^*$, $\bt^*$) using robust registration algorithm (e.g. RANSAC) on $\cI^*(\cJ)$.
\end{algorithmic}
\label{alg:MainAlgorithm}
\end{algorithm}

\vspace{-0mm}
\section{Experiments}
\label{sec:experiments}
\vspace{-0mm}
In this section, we test our proposed method on both synthetic and real-world dataset. All experiments are executed on Intel Core 3.70GHz i7 CPU with 32Gb RAM, and Geforce GTX 1080Ti GPU.

\textbf{Benchmark Dataset.}
We evaluate the performance of our proposed method using KITTI odometry dataset \cite{kitti}. This dataset contains point clouds captured from a moving car, equipped with a Velodyne HDL64 LiDAR, around the city of Karlsruhe, Germany. The original point clouds contains around 120,000 points. Since the ground truth error in the original KITTI odometry dataset is large, we utilize the ground truth poses from Semantic KITTI \cite{behley2019iccv} instead. 

\textbf{Baseline Algorithms.}
To show the significant performance of our method, we compare it against the following state-of-the-art pipelines or approaches on the same dataset:
\vspace{-0mm}
\begin{itemize}
\item GORE \cite{gore6dof}: Guaranteed Outlier Removal for point cloud registration. The code was provided by the authors.
\item Practical Maximum Clique with Pairwise Constraints (PMC) \cite{parra19pairwise}. The code was provided by the authors.
\item Keypoint-based 4PCS (K4PCS)\cite{THEILER2014149}: A variant of 4PCS, which is applied on key points.
\vspace{-0mm}
\end{itemize}

We down-sample the input point clouds using a voxel size of 0.1m. We then apply Intrinsic Shape Signature (ISS)\cite{ISS} to detect keypoints and Fast Point Feature Histogram (FPFH)\cite{FPFH} to compute local geometric features of each point. Finally, we generate initial correspondences based on these features using K nearest neighbors (K = 10) while setting the inlier threshold to 0.1. 

\textbf{Evaluation Criteria} 
The evaluation is based on calculating the angular and translational error of the estimated transformation ($R, T$) against the ground truth ($\overline{R}, \overline{T}$). 
The angular error is calculated as $angErr = 2\arcsin{\frac{\| R - \overline{R}\|}{2\sqrt{2}}}$.
The translational error is calculated by the Euclidean distance between $T$ and $\overline{T}$. 
\vspace{-0mm}
\subsection{Results on the Semi-synthetic Data}
\vspace{-0mm}
We first show the performance of our method on semi-synthetically generated data. The point cloud ``000001.bin" from sequence 00 of KITTI dataset is loaded and considered as source point cloud $M$. In this experiment, we use ground-truth semantic information. Motivated by \cite{SDRSAC}, to produce target point cloud $N$, we apply a random 6DOF transformation and a small Gaussian noise (with zero mean and variance of 0.01) to $N$. We randomly select and remove $k$\% points in $N$ to simulate overlapping rate. We conduct the experiments with k = 10\%, 15\%,..., 50\%. 

\begin{figure}[h]
    \centering
    \includegraphics[width=1\linewidth]{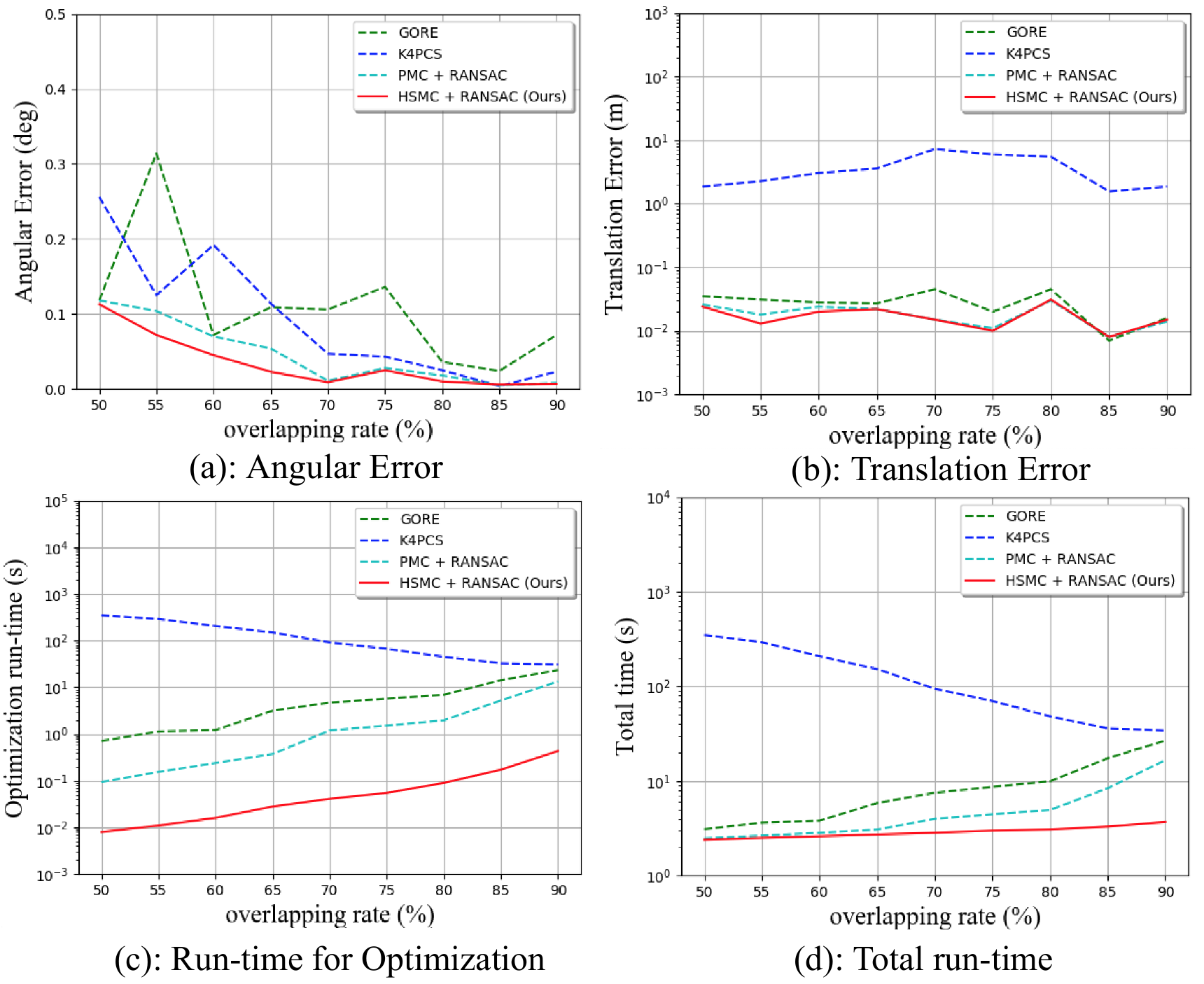}
    \caption{Performance of the proposed method on semi-synthetic data in comparison to other methods.}
    \label{fig:SynPerformance}
    \vspace{-3mm}
\end{figure}

Fig.~\ref{fig:SynPerformance} shows the median values (angular error, translation error, optimization run-time, and total run-time) in 100 runs for all methods. In this stage, as we use the ground-truth for semantic information, the total time does not include the run-time for semantic predictions. (The semantic prediction run-time {\em is} reported in section \ref{SubSecReal}).
As shown in the Fig.~\ref{fig:SynPerformance}, it is evident that our algorithm outperforms the above methods in term of run-time with better (or comparable) errors to the others. The performance of K4PCS is unstable because it is sensitive to input parameters such as approximate overlapping, etc.

\begin{table}
\vspace{-5mm}
\caption{Number of putative correspondences before and after semantic pruning step, and the number of inliers.}
\label{tab:CompareCorr}
\centering
\scalebox{0.82}{
\begin{tabular}{|l|r;{1pt/1pt}r;{1pt/1pt}r;{1pt/1pt}r;{1pt/1pt}r;{1pt/1pt}r;{1pt/1pt}r;{1pt/1pt}r;{1pt/1pt}r|} 
\hline
Overlapping rate (\%)                              & 50   & 55   & 60   & 65   & 70   & 75   & 80   & 85   & 90    \\ 
\hline
Putative Correspondences (Before semantic pruning) & 2384 & 2899 & 3393 & 4023 & 4728 & 5162 & 5800 & 6402 & 7041  \\
Putative Correspondences (After semantic pruning)  & 809  & 1054 & 1274 & 1735 & 2087 & 2427 & 2771 & 3416 & 4027  \\
Number of inliers                                  & 39   & 46   & 73   & 109  & 164  & 193  & 287  & 362  & 495   \\
\hline
\end{tabular}
}
\end{table}
To evaluate the robustness of our method, we add noise to the semantic labels in the target point cloud. In particular, we randomly select and change $h$\% point labels. We conduct the experiments with h = 10\%, 20\%,..., 90\%. Table~\ref{tab:SemanticNoise} shows the median values (angular error, translation error, and optimization run-time) in 100 runs for our method. As shown in the table, our method is robust to  noise in semantic label. We report the results with the noise rate up to 70\%. Our method fails when the noise rate is over 80\%. However, most of the recent deep learning networks \cite{wu2017squeezeseg,wu2018squeezesegv2,Darknet53} for point cloud semantic segmentation can achieve over 70\% average accuracy. As a result, our pipeline is flexible such that practitioners can use any deep learning network suited to their application.
\vspace{-2mm}
\begin{table}
\vspace{-5mm}
\caption{Accuracy and run-time of our pipeline when adding noise to semantic segmentation.}
\label{tab:SemanticNoise}
\centering
\scalebox{0.85}{
\begin{tabular}{|l|l|r;{1pt/1pt}r;{1pt/1pt}r;{1pt/1pt}r;{1pt/1pt}r;{1pt/1pt}r;{1pt/1pt}r|} 
\hline
             & Noise rate (\% & \multicolumn{1}{c;{1pt/1pt}}{10} & \multicolumn{1}{c;{1pt/1pt}}{20} & \multicolumn{1}{c;{1pt/1pt}}{30} & \multicolumn{1}{c;{1pt/1pt}}{40} & \multicolumn{1}{c;{1pt/1pt}}{50} & \multicolumn{1}{c;{1pt/1pt}}{60} & \multicolumn{1}{c|}{70}  \\ 
\hline
HSMC + RANSAC & AngErr (deg)   & ~0.007                           & ~0.011                           & ~0.013                           & ~0.036                           & ~0.124                           & ~0.279                           & ~0.166                   \\
             & TrErr (m)      & 0.015                            & 0.013                            & 0.026                            & 0.018                            & 0.037                            & 0.037                            & 0.017                    \\
             & Run-time (s)   & 0.149                            & 0.046                            & 0.025                            & 0.011                            & 0.007                            & 0.005                            & 0.005                    \\
\hline
\end{tabular}
}
\label{tab:AccuracySemantic}
\end{table}

\subsection{Results on the Real Data}\label{SubSecReal}
To demonstrate the practicality of our method, we evaluate the performance on the KITTI odometry benchmark. Based on the benchmark, we use 11 sequences (from 11 to 21) for testing. Motivated by DeepVCP \cite{DeepVCP} experiment setting, we sample the input source from every 150 frame intervals and register with all target scans within 4m translation. 
To extract semantic information, we utilize predictions from a state-of-the-art deep learning model RangeNet53++ \cite{Darknet53}. The mean Intersection over Union (mIoU) and average accuracy obtained are 0.52 and 0.89 respectively.

\begin{figure}[h]
    \centering
    \includegraphics[width=1\linewidth]{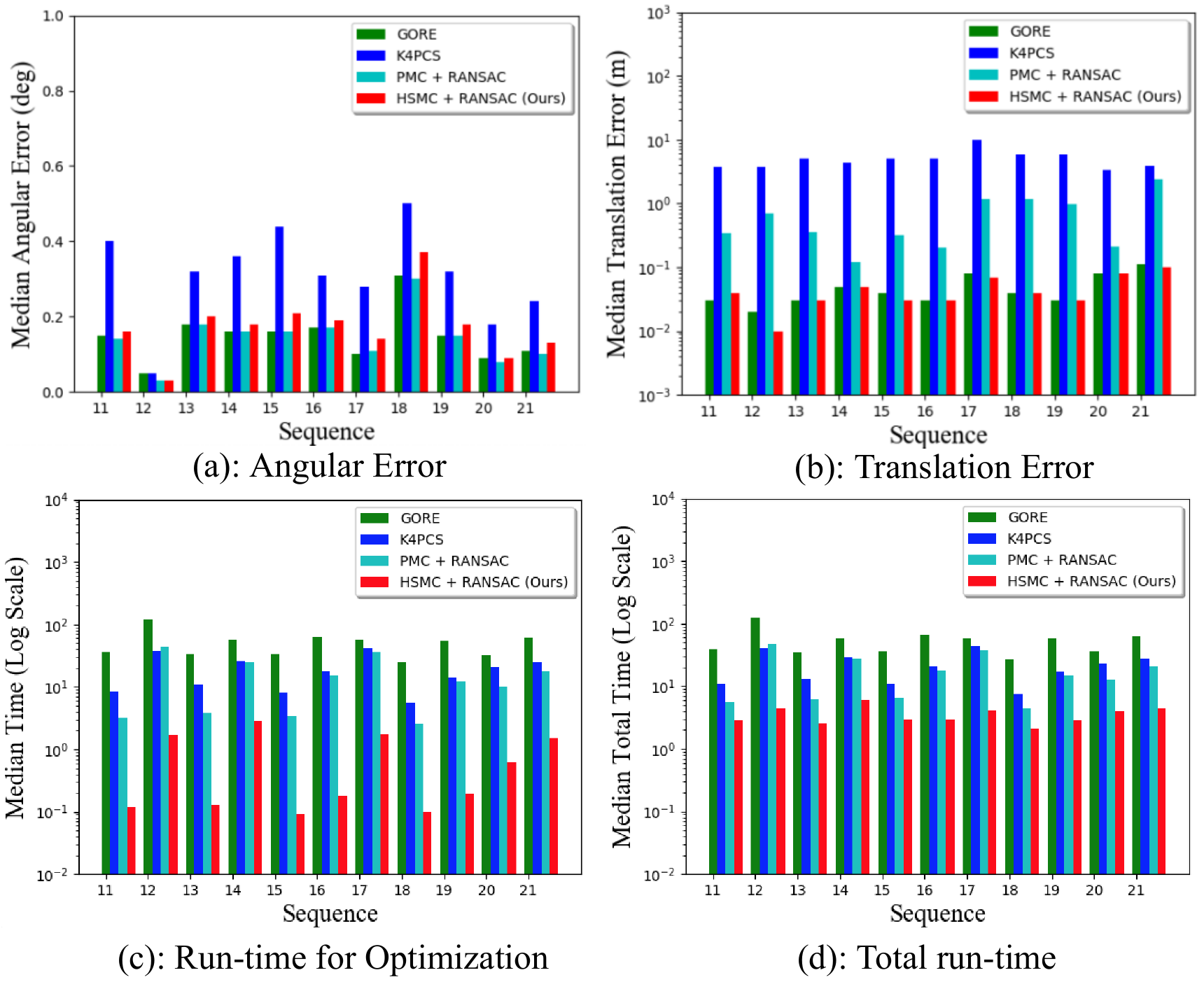}
    \caption{Performance of the proposed method on the KITTI dataset in comparison to other methods.}
    \label{fig:PerformanceKITTI}
\end{figure}

Fig.~\ref{fig:PerformanceKITTI} reports the angular error, translation error, optimization run-time and total run-time of all methods. Note: there are few cases where some methods have failed, which has impacted on the mean values. To be fair, we report  median values which are less influenced, as well as mean (that is, median, mean, standard deviation) in Table~\ref{tab:ComparisionKITTI} and show the comparison between median values of all methods in Fig.~\ref{fig:PerformanceKITTI}. 
The total run-time includes the run-time for extracting key points, generating initial correspondences, semantic information prediction (for only our methods), and optimization run-time. Note that the run-time for semantic prediction was taken from the original paper RangeNet53++ \cite{Darknet53}.
As shown in Fig.~\ref{fig:PerformanceKITTI}, our methods achieve better (or comparable) accuracy with extremely faster run-time than others. Compared to PMC, the optimization run-time of HSMC is significantly faster (up to 99 times). 
In particular, our pipeline is flexible, practitioners can replace PMC with any maximum clique solver, and RangeNet53++ with any state-of-the-art deep learning model.
\begin{table}
\vspace{-5mm}
\caption{Comparison on KITTI dataset. Our method performance achieve comparable accuracy with much faster run-time.}
\label{tab:ComparisionKITTI}
\centering
\scalebox{0.65}{
\begin{tabular}{|r|l|r;{1pt/1pt}r;{1pt/1pt}r|r;{1pt/1pt}r;{1pt/1pt}r|r;{1pt/1pt}r;{1pt/1pt}r|r;{1pt/1pt}r;{1pt/1pt}r|} 
\hline
\multicolumn{1}{|r}{}           & \multicolumn{1}{r|}{} & \multicolumn{3}{c|}{GORE}                                                                            & \multicolumn{3}{c|}{K4PCS}                                                                           & \multicolumn{3}{c|}{PMC + RANSAC}                                                                    & \multicolumn{3}{c|}{HSMC + RANSAC}                                                                    \\
\multicolumn{1}{|l}{}           &                       & \multicolumn{3}{l|}{}                                                                                & \multicolumn{3}{l|}{}                                                                                & \multicolumn{3}{l|}{}                                                                                & \multicolumn{3}{c|}{(Ours)}                                                                           \\ 
\cline{3-14}
\multicolumn{1}{|r}{}           & \multicolumn{1}{r|}{} & \multicolumn{1}{c;{1pt/1pt}}{Median} & \multicolumn{1}{c;{1pt/1pt}}{Mean} & \multicolumn{1}{c|}{Std} & \multicolumn{1}{c;{1pt/1pt}}{Median} & \multicolumn{1}{c;{1pt/1pt}}{Mean} & \multicolumn{1}{c|}{Std} & \multicolumn{1}{c;{1pt/1pt}}{Median} & \multicolumn{1}{c;{1pt/1pt}}{Mean} & \multicolumn{1}{c|}{Std} & \multicolumn{1}{c;{1pt/1pt}}{Median} & \multicolumn{1}{c;{1pt/1pt}}{Mean} & \multicolumn{1}{c|}{Std}  \\ 
\hline
\multicolumn{1}{|l|}{Seq.11}    & angErr (deg)          & 0.15                                 & \textbf{0.17}                      & \textbf{0.13}            & 0.40                                 & 6.64                               & 32.59                    & \textbf{0.14}                        & \textbf{0.17}                      & 0.14                     & 0.16                                 & 0.21                               & 0.17                      \\
\multicolumn{1}{|l|}{58 pairs}  & trErr (m)             & \textbf{0.03}                        & 0.09                               & 0.39                     & 3.79                                 & 6.38                               & 7.09                     & 0.34                                 & 0.49                               & 0.54                     & 0.04                                 & \textbf{0.04}                      & \textbf{0.02}             \\
                                & run-time (s)          & 36.52                                & 42.69                              & 27.99                    & 8.57                                 & 11.98                              & 10.10                    & 3.21                                 & 4.10                               & 3.49                     & \textbf{0.12}                        & \textbf{0.15}                      & \textbf{0.11}             \\
                                & total time (s)        & 38.83                                & 45.11                              & 27.93                    & 10.93                                & 14.39                              & 10.09                    & 5.63                                 & 6.51                               & 3.43                     & \textbf{2.83}                        & \textbf{2.74}                      & \textbf{0.28}             \\
                                & \multicolumn{1}{r|}{} &                                      &                                    &                          &                                      &                                    &                          &                                      &                                    &                          &                                      &                                    &                           \\ 
\cdashline{1-1}[1pt/1pt]\cline{2-14}
\multicolumn{1}{|l|}{Seq. 12}   & angErr (deg)          & 0.05                                 & 0.12                               & 0.15                     & 0.05                                 & 0.18                               & 0.23                     & \textbf{0.03}                        & \textbf{0.08}                      & \textbf{0.12}            & \textbf{0.03}                        & 0.33                               & 1.48                      \\
\multicolumn{1}{|l|}{46 pairs}  & trErr (m)             & 0.02                                 & 0.59                               & 1.16                     & 3.78                                 & 4.57                               & 2.50                     & 0.71                                 & 1.34                               & 1.51                     & \textbf{0.01}                        & \textbf{0.51}                      & \textbf{1.07}             \\
                                & run-time (s)          & 121.55                               & 116.45                             & 51.82                    & 37.72                                & 37                                 & 12.99                    & 43.63                                & 48.52                              & 36.16                    & \textbf{1.71}                        & \textbf{1.64}                      & \textbf{0.73}             \\
                                & total time (s)        & 123.99                               & 119.27                             & 51.39                    & 40.25                                & 39.82                              & 12.54                    & 46.17                                & 51.34                              & 35.82                    & \textbf{4.52}                        & \textbf{4.64}                      & \textbf{0.53}             \\
                                & \multicolumn{1}{r|}{} &                                      &                                    &                          &                                      &                                    &                          &                                      &                                    &                          &                                      &                                    &                           \\ 
\hline
\multicolumn{1}{|l|}{Seq. 13}   & angErr (deg)          & \textbf{0.18}                        & \textbf{0.23}                      & \textbf{0.20}            & 0.32                                 & 1.27                               & 12.16                    & \textbf{0.18}                        & 0.26                               & 0.23                     & 0.20                                 & 0.28                               & 0.28                      \\
\multicolumn{1}{|l|}{209 pairs} & trErr (m)             & \textbf{0.03}                        & 0.06                               & 0.27                     & 4.99                                 & 6.16                               & 4.59                     & 0.35                                 & 0.55                               & 0.57                     & \textbf{0.03}                        & \textbf{0.04}                      & \textbf{0.04}             \\
                                & run-time (s)          & 33.21                                & 41.80                              & 30.51                    & 10.92                                & 11.32                              & 5.46                     & 3.88                                 & 29.19                              & 150.77                   & \textbf{0.13}                        & \textbf{0.31}                      & \textbf{0.73}             \\
                                & total time (s)        & 35.25                                & 44.00                              & 30.76                    & 13.07                                & 13.53                              & 5.72                     & 6.16                                 & 31.40                              & 150.88                   & \textbf{2.50}                        & \textbf{2.69}                      & \textbf{0.94}             \\
                                & \multicolumn{1}{r|}{} &                                      &                                    &                          &                                      &                                    &                          &                                      &                                    &                          &                                      &                                    &                           \\ 
\hline
\multicolumn{1}{|l|}{Seq. 14}   & angErr (deg)          & \textbf{0.16}                        & \textbf{0.18}                      & \textbf{0.11}            & 0.36                                 & 9.54                               & 37.21                    & \textbf{0.16}                        & 0.21                               & 0.18                     & 0.18                                 & 0.34                               & 0.57                      \\
\multicolumn{1}{|l|}{67 pairs}  & trErr (m)             & \textbf{0.05}                        & 0.11                               & 0.41                     & 4.30                                 & 6.10                               & 4.05                     & 0.12                                 & 0.17                               & 0.14                     & \textbf{0.05}                        & \textbf{0.07}                      & 0.07                      \\
                                & run-time (s)          & 56.80                                & 58.58                              & 23.05                    & 25.53                                & 27.36                              & 11.31                    & 24.49                                & 24.83                              & 7.97                     & \textbf{2.82}                        & \textbf{2.78}                      & \textbf{1.00}             \\
                                & total time (s)        & 59.60                                & 61.89                              & 23.21                    & 29.11                                & 30.67                              & 11.47                    & 27.73                                & 28.14                              & 8.12                     & \textbf{6.11}                        & \textbf{6.27}                      & \textbf{1.27}             \\
                                & \multicolumn{1}{r|}{} &                                      &                                    &                          &                                      &                                    &                          &                                      &                                    &                          &                                      &                                    &                           \\ 
\hline
\multicolumn{1}{|l|}{Seq. 15}   & angErr (deg)          & \textbf{0.16}                        & \textbf{0.20}                      & \textbf{0.14}            & 0.44                                 & 4.48                               & 25.76                    & \textbf{0.16}                        & \textbf{0.20}                      & \textbf{0.14}            & 0.21                                 & 0.29                               & 0.22                      \\
\multicolumn{1}{|l|}{137 pairs} & trErr (m)             & 0.04                                 & 0.11                               & 0.43                     & 5.13                                 & 6.23                               & 5.05                     & 0.32                                 & 0.47                               & 0.45                     & \textbf{0.03}                        & \textbf{0.04}                      & \textbf{0.03}             \\
                                & run-time (s)          & 33.08                                & 38.25                              & 22.40                    & 8.09                                 & 10.06                              & 5.80                     & 3.45                                 & 6.80                               & 7.46                     & \textbf{0.09}                        & \textbf{0.11}                      & \textbf{0.05}             \\
                                & total time (s)        & 36.09                                & 41.04                              & 22.37                    & 10.88                                & 12.85                              & 5.81                     & 6.50                                 & 9.60                               & 7.46                     & \textbf{3.01}                        & \textbf{3.08}                      & \textbf{0.31}             \\
                                & \multicolumn{1}{r|}{} &                                      &                                    &                          &                                      &                                    &                          &                                      &                                    &                          &                                      &                                    &                           \\ 
\hline
\multicolumn{1}{|l|}{Seq. 16}   & angErr (deg)          & \textbf{0.17}                        & \textbf{0.20}                      & \textbf{0.10}            & 0.31                                 & 0.35                               & 0.25                     & \textbf{0.17}                        & 0.21                               & 0.14                     & 0.19                                 & 0.24                               & 0.18                      \\
\multicolumn{1}{|l|}{116 pairs} & trErr (m)             & \textbf{0.03}                        & \textbf{0.03}                      & \textbf{0.02}            & 5.14                                 & 6.54                               & 4.53                     & 0.20                                 & 0.36                               & 0.46                     & \textbf{0.03}                        & \textbf{0.03}                      & \textbf{0.02}             \\
                                & run-time (s)          & 62.27                                & 68.62                              & 36.78                    & 17.84                                & 18.29                              & 7.36                     & 15.46                                & 19.46                              & 18.80                    & \textbf{0.18}                        & \textbf{0.18}                      & \textbf{0.06}             \\
                                & total time (s)        & 64.90                                & 71.21                              & 36.74                    & 20.50                                & 20.88                              & 7.31                     & 18.05                                & 22.05                              & 18.76                    & \textbf{2.95}                        & \textbf{2.95}                      & \textbf{0.16}             \\
                                & \multicolumn{1}{r|}{} &                                      &                                    &                          &                                      &                                    &                          &                                      &                                    &                          &                                      &                                    &                           \\ 
\hline
\multicolumn{1}{|l|}{Seq. 17}   & angErr (deg)          & \textbf{0.10}                        & \textbf{0.11}                      & \textbf{0.04}            & 0.28                                 & 0.68                               & 1.3                      & 0.11                                 & 0.12                               & 0.06                     & 0.14                                 & 0.17                               & 0.10                      \\
\multicolumn{1}{|l|}{14 pairs}  & trErr (m)             & 0.08                                 & \textbf{1.04}                      & 1.35                     & 9.86                                 & 9.05                               & 4.73                     & 1.18                                 & 1.69                               & 1.64                     & \textbf{0.07}                        & \textbf{1.04}                      & \textbf{1.34}             \\
                                & run-time (s)          & 57.34                                & 59.36                              & 17.45                    & 41.94                                & 40.68                              & 8.13                     & 35.61                                & 32.44                              & 7.16                     & \textbf{1.72}                        & \textbf{2.15}                      & \textbf{1.06}             \\
                                & total time (s)        & 59.34                                & 61.57                              & 17.60                    & 43.94                                & 42.89                              & 7.76                     & 37.37                                & 34.65                              & 6.66                     & \textbf{4.10}                        & \textbf{4.54}                      & \textbf{1.16}             \\
                                & \multicolumn{1}{r|}{} &                                      &                                    &                          &                                      &                                    &                          &                                      &                                    &                          &                                      &                                    &                           \\ 
\hline
\multicolumn{1}{|l|}{Seq.18}    & angErr (deg)          & 0.31                                 & \textbf{0.42}                      & \textbf{0.33}            & 0.50                                 & 2.13                               & 15.75                    & \textbf{0.30}                        & 0.45                               & 0.37                     & 0.37                                 & 0.52                               & 0.39                      \\
\multicolumn{1}{|l|}{121 pairs} & trErr (m)             & \textbf{0.04}                        & \textbf{0.04}                      & \textbf{0.02}            & 5.94                                 & 6.66                               & 4.22                     & 1.17                                 & 1.57                               & 1.52                     & \textbf{0.04}                        & 0.05                               & 0.04                      \\
                                & run-time (s)          & 24.98                                & 32.82                              & 24.37                    & 5.62                                 & 7.65                               & 4.97                     & 2.54                                 & 26.28                              & 169.89                   & \textbf{0.10}                        & \textbf{0.49}                      & \textbf{1.50}             \\
                                & total time (s)        & 26.62                                & 34.63                              & 24.49                    & 7.42                                 & 9.46                               & 5.23                     & 4.41                                 & 28.10                              & 169.93                   & \textbf{2.12}                        & \textbf{2.48}                      & \textbf{1.61}             \\
                                & \multicolumn{1}{r|}{} &                                      &                                    &                          &                                      &                                    &                          &                                      &                                    &                          &                                      &                                    &                           \\ 
\hline
\multicolumn{1}{|l|}{Seq. 19}   & angErr (deg)          & \textbf{0.15}                        & \textbf{0.18}                      & \textbf{0.12}            & 0.32                                 & 2.01                               & 16.67                    & \textbf{0.15}                        & 0.20                               & 0.19                     & 0.18                                 & 0.24                               & 0.22                      \\
\multicolumn{1}{|l|}{340 pairs} & trErr (m)             & \textbf{0.03}                        & \textbf{0.03}                      & \textbf{0.02}            & 5.78                                 & 6.58                               & 4.50                     & 0.96                                 & 1.51                               & 2.41                     & \textbf{0.03}                        & \textbf{0.03}                      & \textbf{0.02}             \\
                                & run-time (s)          & 55.39                                & 61.21                              & 36.90                    & 14.20                                & 16.41                              & 9.14                     & 12.02                                & 23.19                              & 48.57                    & \textbf{0.19}                        & \textbf{0.32}                      & \textbf{0.83}             \\
                                & total time (s)        & 58.00                                & 63.64                              & 36.97                    & 16.88                                & 18.83                              & 9.19                     & 14.62                                & 25.61                              & 48.56                    & \textbf{2.88}                        & \textbf{2.92}                      & \textbf{0.91}             \\
                                & \multicolumn{1}{r|}{} &                                      &                                    &                          &                                      &                                    &                          &                                      &                                    &                          &                                      &                                    &                           \\ 
\hline
\multicolumn{1}{|l|}{Seq. 20}   & angErr (deg)          & 0.09                                 & \textbf{0.11}                      & \textbf{0.08}            & 0.18                                 & 0.24                               & 0.29                     & \textbf{0.08}                        & 0.12                               & 0.17                     & 0.09                                 & 0.16                               & 0.19                      \\
\multicolumn{1}{|l|}{63 pairs}  & trErr (m)             & \textbf{0.08}                        & 0.16                               & \textbf{0.20}            & 3.32                                 & 3.54                               & 2.66                     & 0.21                                 & 0.53                               & 0.67                     & \textbf{0.08}                        & \textbf{0.15}                      & \textbf{0.20}             \\
                                & run-time (s)          & 32.95                                & 40.28                              & 23.27                    & 20.55                                & 19.77                              & 7.09                     & 10.18                                & 11.52                              & 6.38                     & \textbf{0.61}                        & \textbf{0.80}                      & \textbf{0.46}             \\
                                & total time (s)        & 36.11                                & 43.16                              & 23.31                    & 23.27                                & 22.66                              & 7.14                     & 12.90                                & 14.41                              & 6.38                     & \textbf{3.99}                        & \textbf{3.86}                      & \textbf{0.39}             \\
                                & \multicolumn{1}{r|}{} &                                      &                                    &                          &                                      &                                    &                          &                                      &                                    &                          &                                      &                                    &                           \\ 
\hline
\multicolumn{1}{|l|}{Seq. 21}   & angErr (deg)          & 0.11                                 & \textbf{0.13}                      & \textbf{0.09}            & 0.24                                 & 0.30                               & 0.21                     & \textbf{0.10}                        & 0.14                               & 0.11                     & 0.13                                 & 0.19                               & 0.20                      \\
\multicolumn{1}{|l|}{40 pairs}  & trErr (m)             & 0.11                                 & 0.54                               & 0.85                     & 3.92                                 & 4.48                               & 3.16                     & 2.36                                 & 4.25                               & 4.80                     & \textbf{0.10}                        & \textbf{0.39}                      & \textbf{0.72}             \\
                                & run-time (s)          & 60.96                                & 64.85                              & 26.24                    & 25.21                                & 29.06                              & 10.02                    & 17.69                                & 19.38                              & 5.38                     & \textbf{1.50}                        & \textbf{1.53}                      & \textbf{0.64}             \\
                                & total time (s)        & 63.46                                & 67.71                              & 26.17                    & 28.04                                & 31.93                              & 10.07                    & 20.35                                & 22.24                              & 5.39                     & \textbf{4.48}                        & \textbf{4.57}                      & \textbf{0.78}             \\
\multicolumn{1}{|l|}{}          &                       & \multicolumn{1}{l;{1pt/1pt}}{}       & \multicolumn{1}{l;{1pt/1pt}}{}     & \multicolumn{1}{l|}{}    & \multicolumn{1}{l;{1pt/1pt}}{}       & \multicolumn{1}{l;{1pt/1pt}}{}     & \multicolumn{1}{l|}{}    & \multicolumn{1}{l;{1pt/1pt}}{}       & \multicolumn{1}{l;{1pt/1pt}}{}     & \multicolumn{1}{l|}{}    & \multicolumn{1}{l;{1pt/1pt}}{}       & \multicolumn{1}{l;{1pt/1pt}}{}     & \multicolumn{1}{l|}{}     \\
\hline
\end{tabular}
}
\end{table}
\begin{figure}[t]
\centering
\begin{subfigure}{.3\textwidth}
  \centering
  \includegraphics[width=\linewidth]{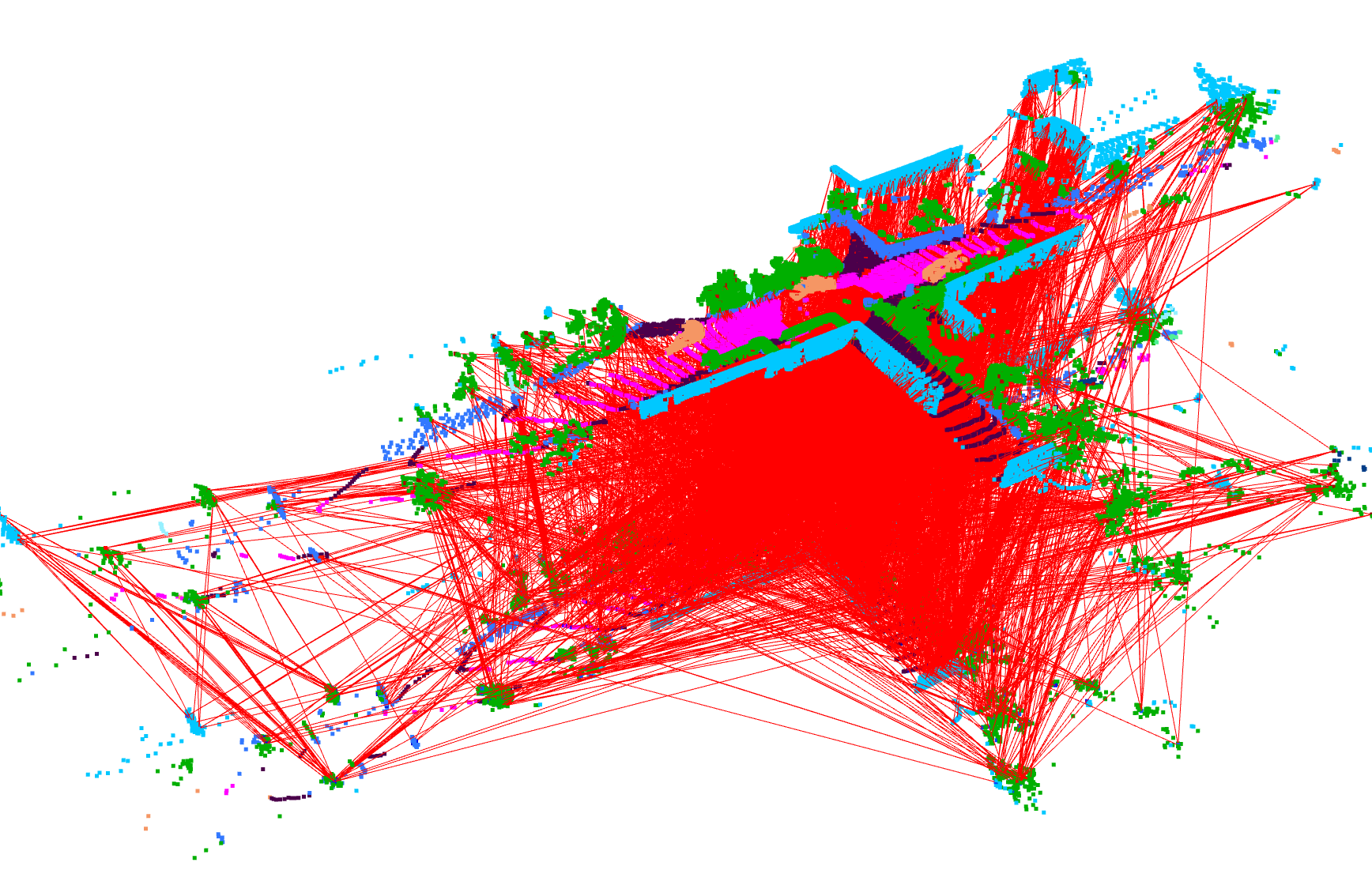}
  \includegraphics[width=\linewidth]{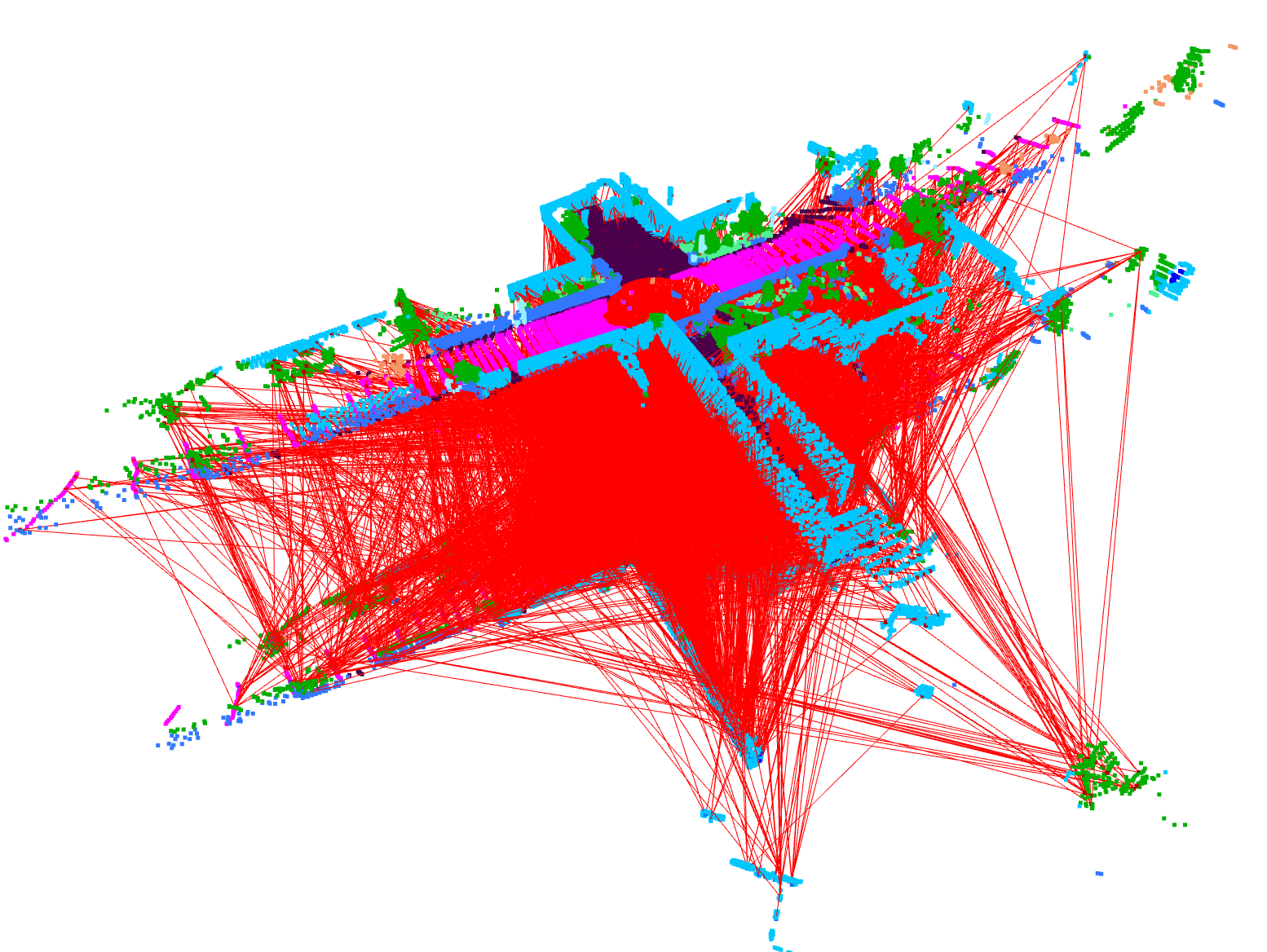}
  \includegraphics[width=\linewidth]{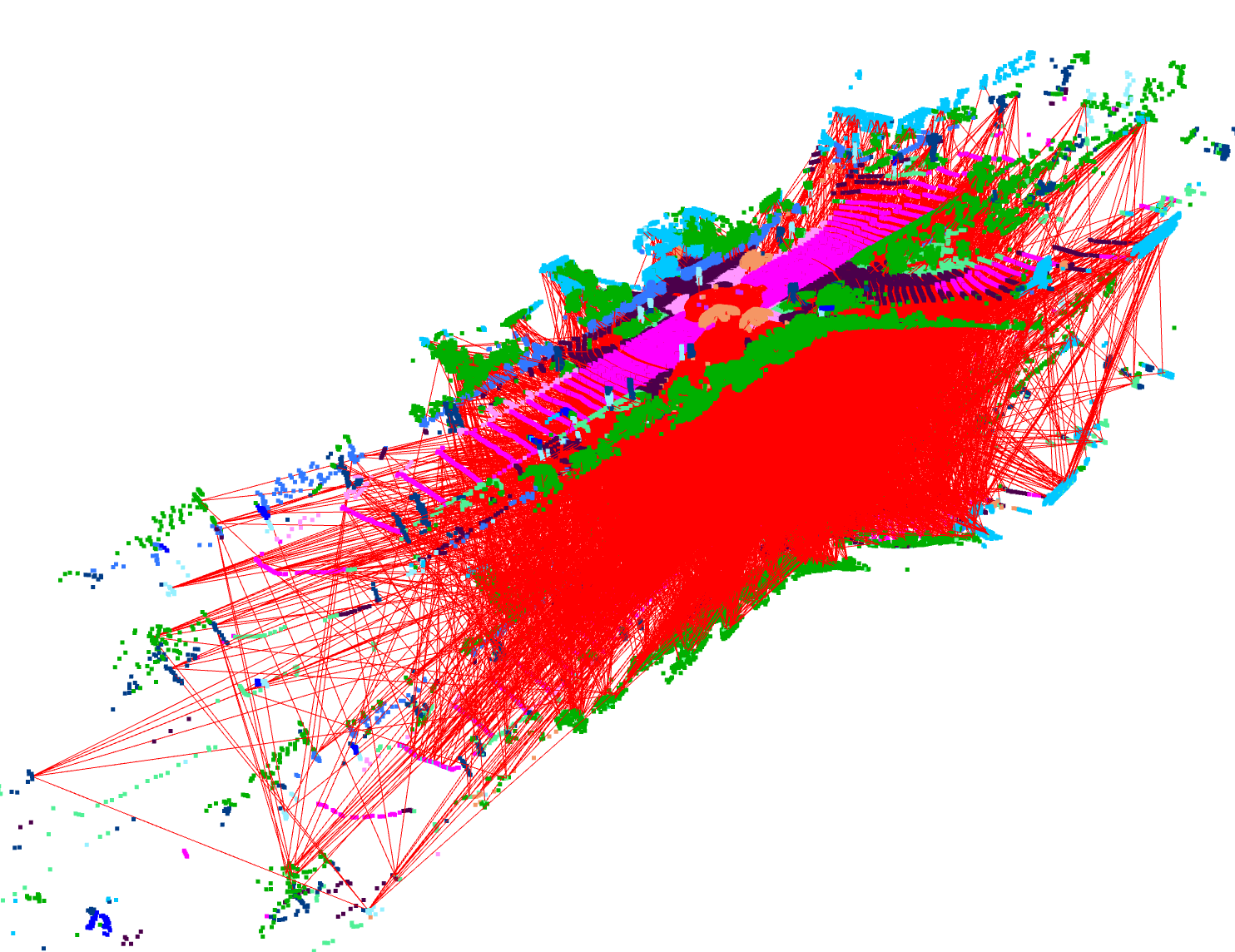}
  \caption{Initial match set}
  \label{fig:sfig1}
\end{subfigure}%
\begin{subfigure}{.3\textwidth}
  \centering
  \includegraphics[width=\linewidth]{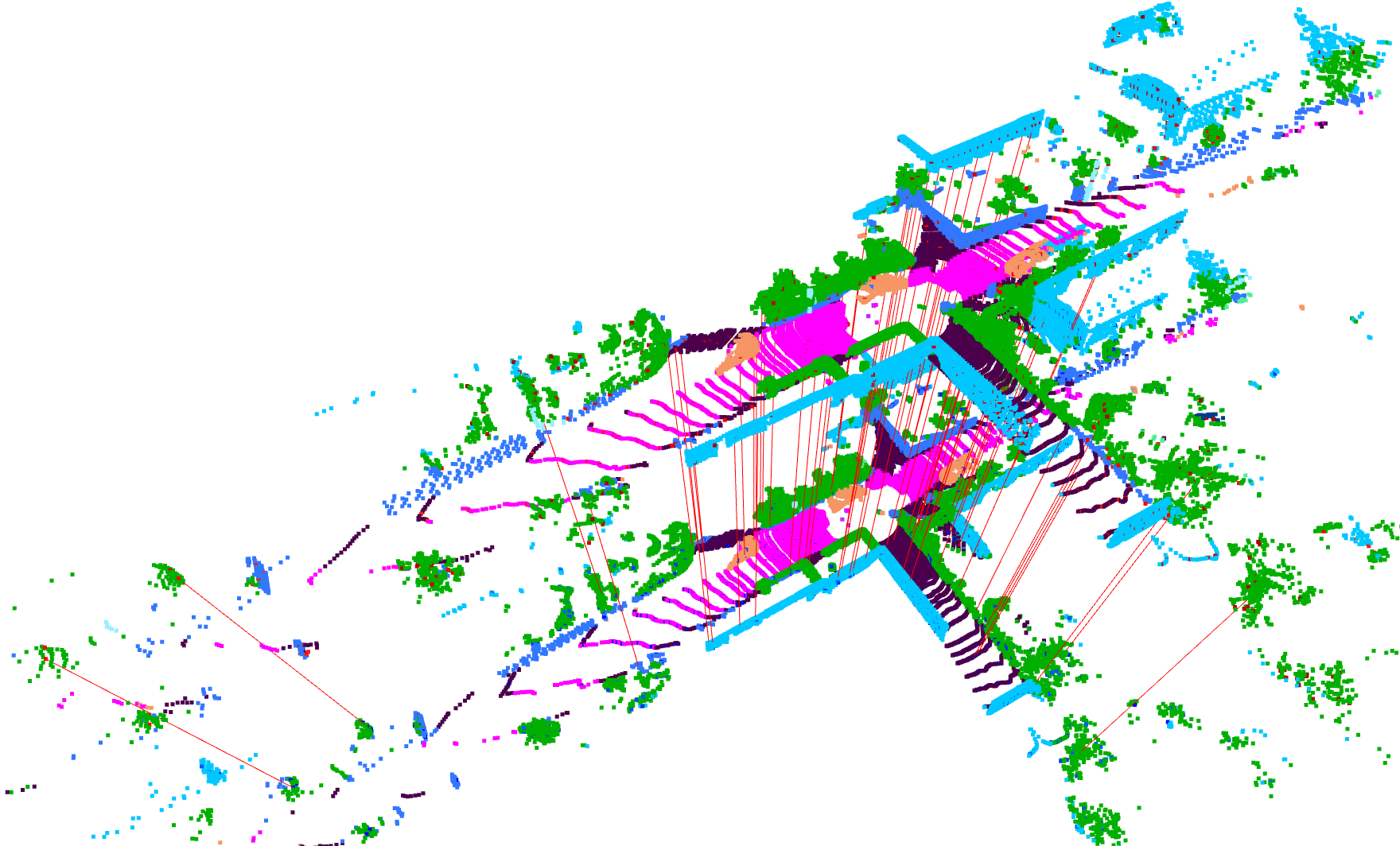}
  \includegraphics[width=\linewidth]{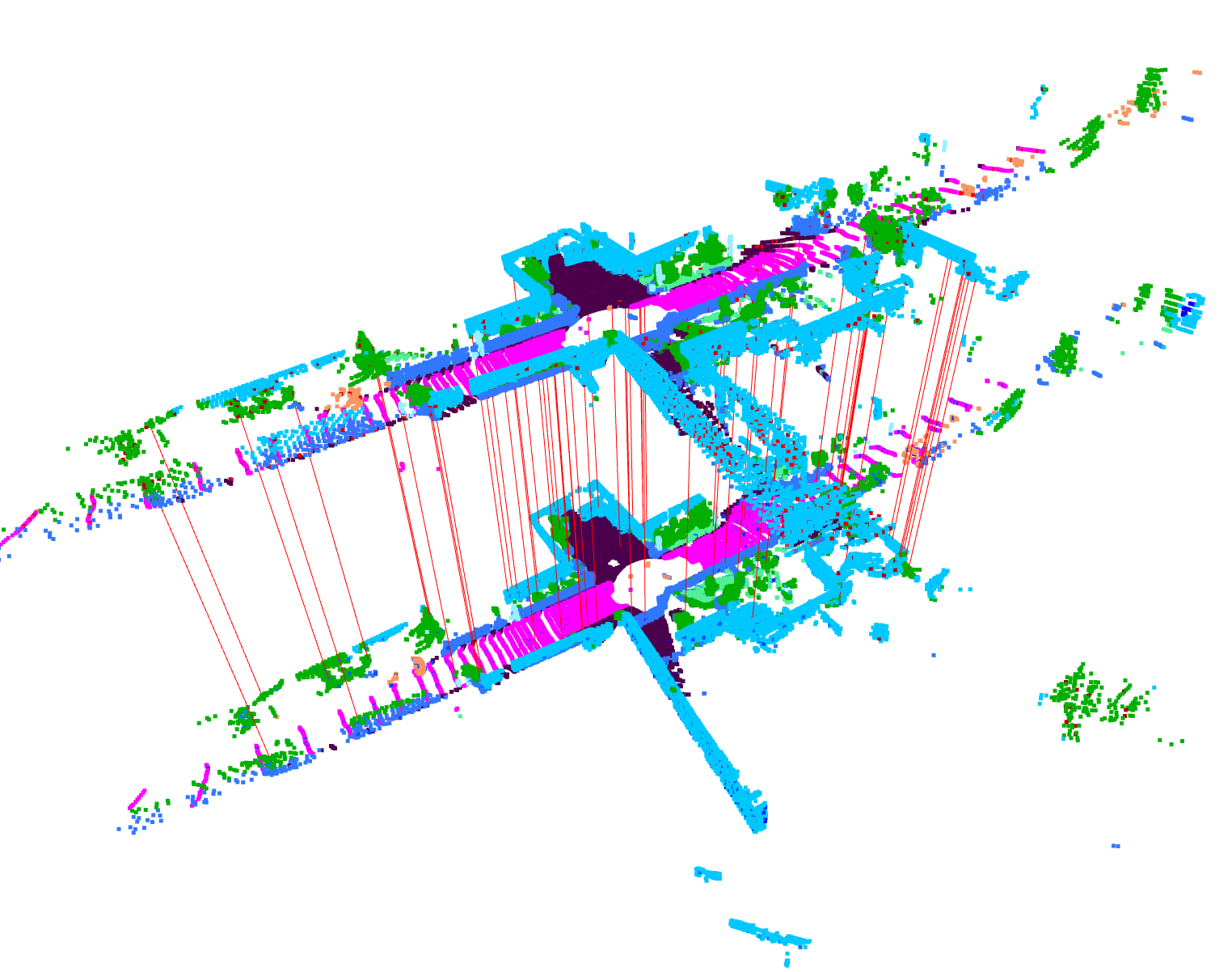}
  \includegraphics[width=\linewidth]{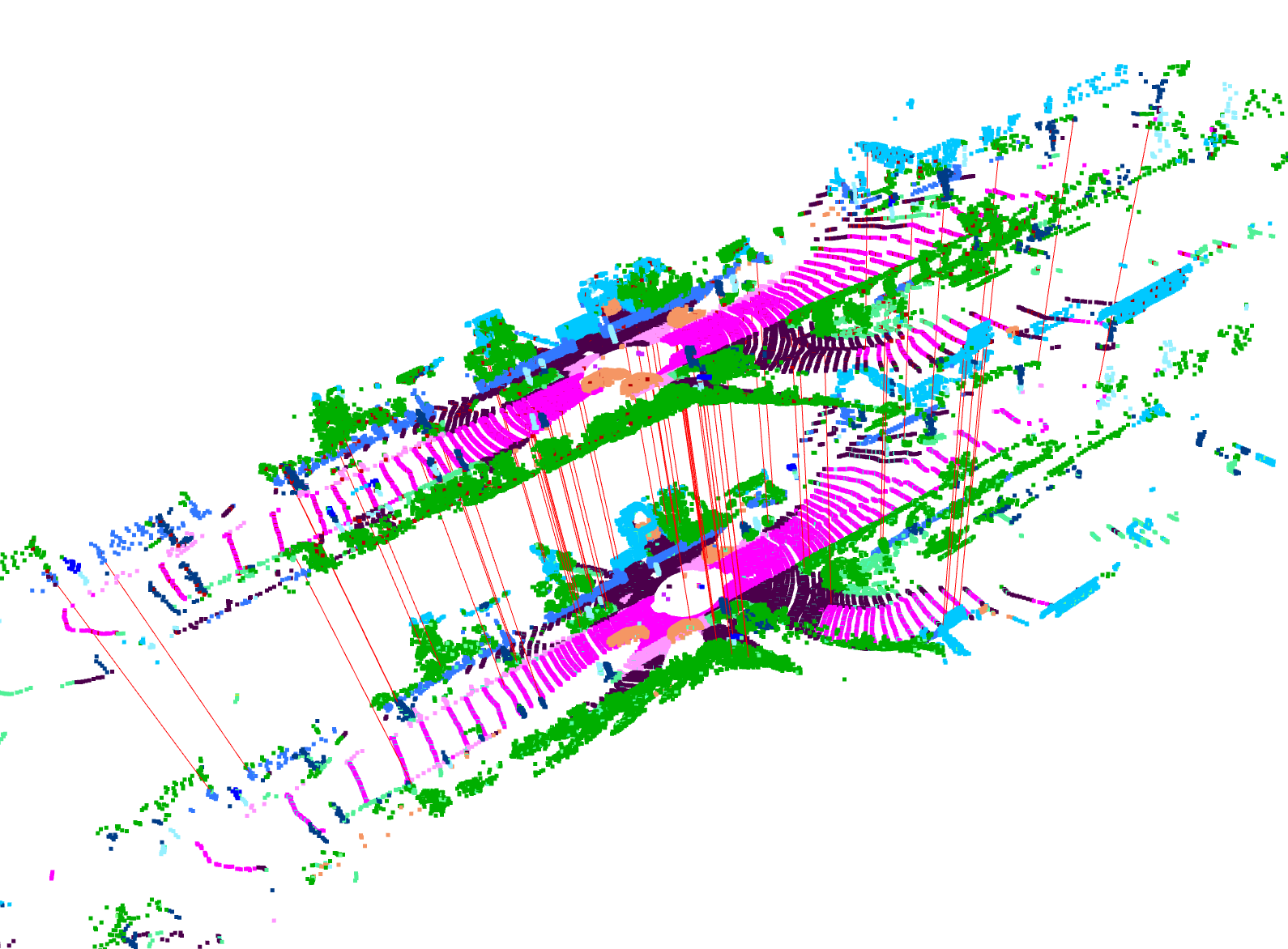}
  \caption{Maximum clique}
  \label{fig:sfig2}
\end{subfigure}
\begin{subfigure}{.3\textwidth}
  \centering
  \includegraphics[width=\linewidth]{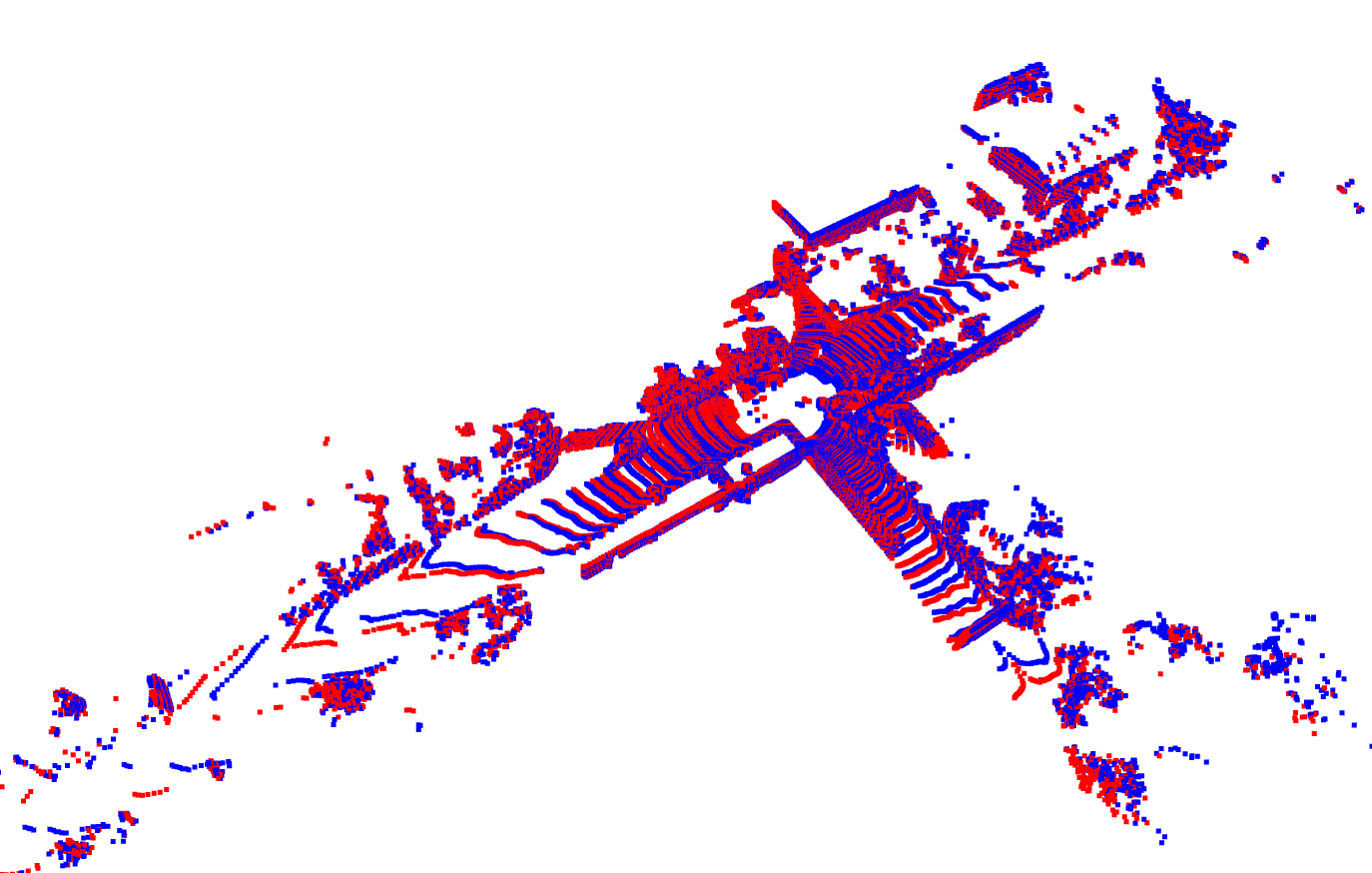}
  \includegraphics[width=\linewidth]{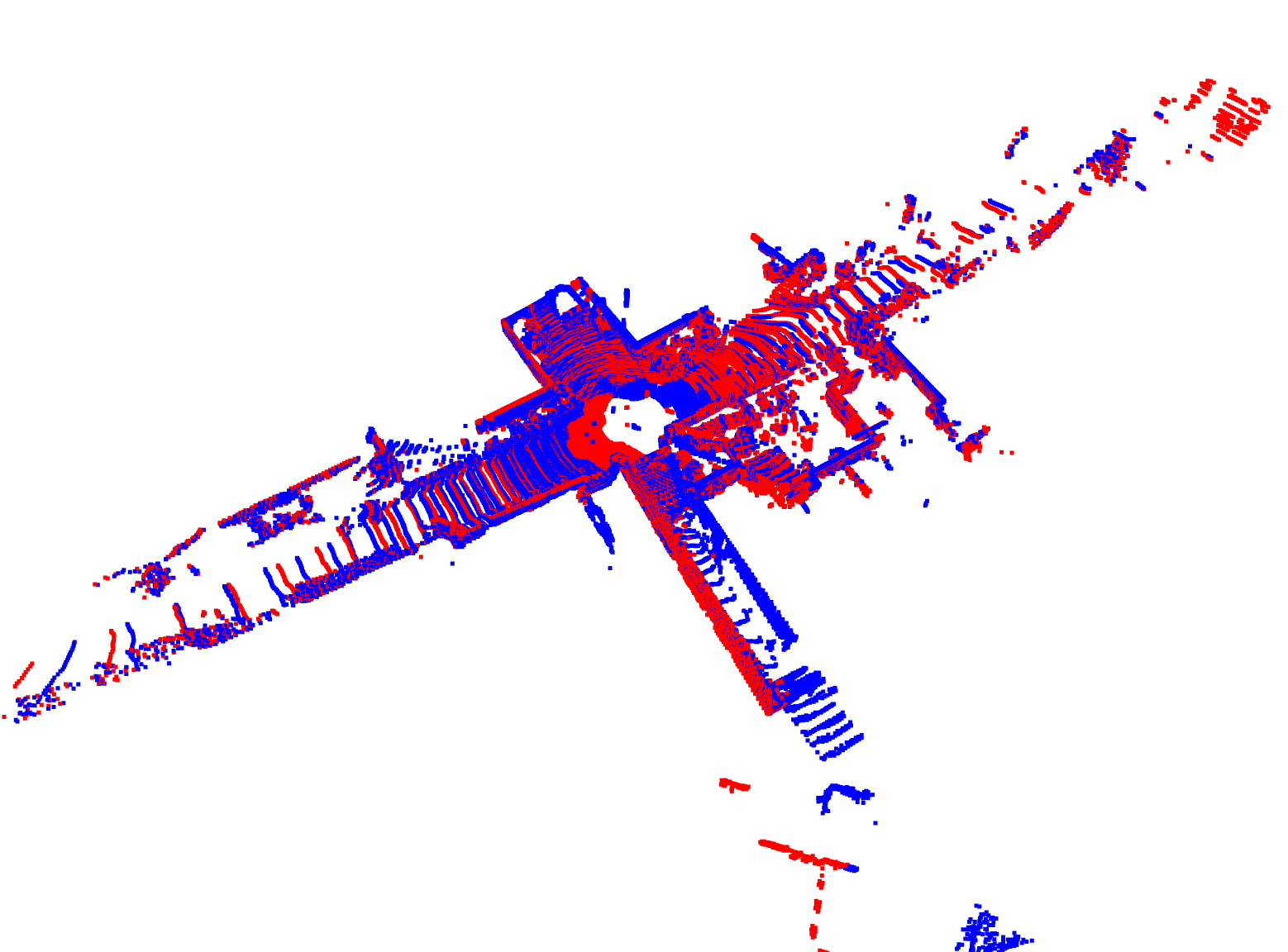}
  \includegraphics[width=\linewidth]{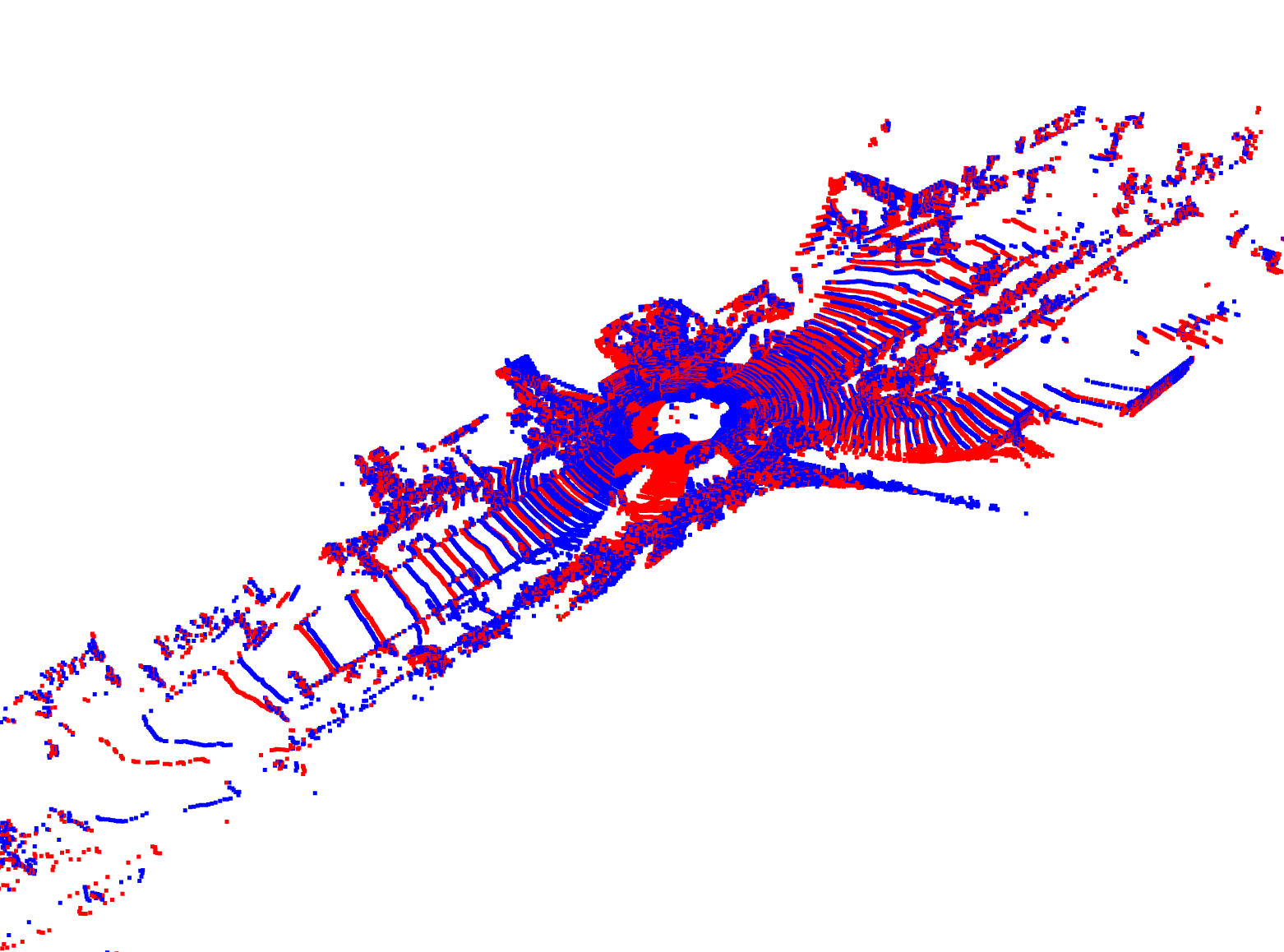}
  \caption{Alignment}
  \label{fig:sfig1}
\end{subfigure}
\caption{Examples visualization of our results.}
\label{fig:fig}
\end{figure}

\section{Conclusion}
We have demonstrated that one can employ semantic constraints, in addition to pairwise conserved distance constraints, to reduce the number of putative matches (remove outliers in those) sufficiently well to make a post-processing step (by RANSAC or other standard technique) computationally attractive even on large data sets. 


\bibliographystyle{splncs}
\bibliography{bibmain}
\end{document}